%% file: main.tex
\documentclass[conference,compsoc]{IEEEtran}

\usepackage[pdftex]{graphicx}
\usepackage{caption}
\usepackage{subfigure}
\usepackage{multirow}
\usepackage{booktabs}
\usepackage{bigstrut}
\usepackage{hyperref}
\usepackage[marginal]{footmisc}
\usepackage[table]{xcolor}
\usepackage{diagbox}
\ifCLASSOPTIONcompsoc
  \usepackage[nocompress]{cite}
\else
  \usepackage{cite}
\fi
\ifCLASSINFOpdf
\else
\fi
\begin{document}

\title{Multi-Semantic Image Recognition Model and Evaluating Index\\
for explaining the deep learning models}

\DeclareRobustCommand*{\IEEEauthorrefmark}[1]{%
    \raisebox{0pt}[0pt][0pt]{\textsuperscript{\footnotesize\ensuremath{#1}}}}
    
\author{\IEEEauthorblockN{Qianmengke Zhao\IEEEauthorrefmark{1},
Ye Wang\IEEEauthorrefmark{2}, Qun Liu\IEEEauthorrefmark{3}\IEEEauthorrefmark{*}}

\IEEEauthorblockA{Chongqing Key Laboratory of Computational Intelligence\\
Chongqing University of Posts and Telecommunications\\
Chongqing, China\\
Email: \IEEEauthorrefmark{1}S200231185@stu.cqupt.edu.cn,
\IEEEauthorrefmark{2}wangye@cqupt.edu.cn,
\IEEEauthorrefmark{3}liuqun@cqupt.edu.cn}}


\maketitle

\renewcommand{\footnoterule}{%
\noindent \rule[0.25\baselineskip]{245pt}{1pt}
}

\def\IEEEkeywordsname{Keywords}
\begin{abstract}
Although deep learning models are powerful among various applications, most deep learning models are still a black box, lacking verifiability and interpretability, which means the decision-making process that human beings cannot understand. Therefore, how to evaluate deep neural networks with explanations is still an urgent task. In this paper, we first propose a multi-semantic image recognition model, which enables human beings to understand the decision-making process of the neural network. Then, we presents a new evaluation index, which can quantitatively assess the model interpretability. We also comprehensively summarize the semantic information that affects the image classification results in the judgment process of neural networks. Finally, this paper also exhibits the relevant baseline performance with current state-of-the-art deep learning models.\footnote{The work is partially supported by NSFC Grant 61936001, and in part by the Science and Technology Research Program of Chongqing Municipal Education Commission under Grants KJQN202100629 and KJQN202100627.}
\end{abstract}


% \fi
\IEEEpeerreviewmaketitle

\section{Introduction}
Driven by deep neural networks, deep learning has made a breakthrough in natural language processing\cite{wang2017comparisons,wang2019icassp,WANG2020340}, image processing\cite{wang2017combining}, speech recognition, and other related fields. In the field of image classification, starting from the AlexNet\cite{krizhevsky2012imagenet} network born in the ImageNet LSVRC-2010 competition in 2012, different neural network models such as VGG\cite{DBLP:journals/corr/SimonyanZ14a}, GoogleNet\cite{he2016deep} and ResNet\cite{2014Going} refresh the evaluation indexes of various competition datasets again and again, and can even achieve better results than human naked eyes' judgment on some datasets.

Although the deep learning model has made remarkable achievements in the evaluation indicators of various datasets, the process of learning features expression is opaque. For humans, the neural network is a "black box model". A large number of parameters in each layer and various nonlinear activation functions map the input to a higher-dimensional space, but higher-dimensional features are hard for humans to understand. Though the output result of the model has high accuracy, it does not accord with the conventional thinking and decision-making process of human beings. Therefore, human beings have a high degree of distrust of neural networks. Especially, in areas such as finance, medical diagnosis, unmanned driving where the decision-making process needs to be clear and logical from input to output, the applicability of the neural network is not very high.

Nowadays, the interpretability of neural networks is an extremely important research direction. Many scholars have put forward their cognition of interpretability. Some scholars use visualization technology based on the gradient interpretation method to explain neural networks. The main methods include deconvolution\cite{DBLP:conf/iccv/ZeilerTF11}, guided backpropagation\cite{simonyan2013deep}, etc. The core idea of this kind of method is to use backpropagation to calculate the visualization results of specific neurons and visually explain what CNN has learned at each layer. The advantage of this kind of method is that it can intuitively detect which features are important through visualization. However, its disadvantages are also obvious: most of the visualization results contain the noise. This method can get important features, but it can not get the contribution of all features to the results. Some scholars study the interpretability independent of the model, among which the typical methods are LIME (Local Interpretable Model-agnostic Explanations)\cite{DBLP:conf/kdd/Ribeiro0G16} and SHAP (Shapley Additive Explanations) \cite{DBLP:conf/nips/LundbergL17}. The idea is to regard the internal structure of the model as a black box and explain the prediction of the model only by analyzing the input and output of the model, that is, try to use a linear model to simulate the original neural network. This method has obvious advantages because it is independent of the internal structure of the model and can be applied to any type of neural network. The disadvantage is whether the linear model can greatly simulate the decision results of the original model, and these methods assume that the semantic features are independent of each other, which is not in line with reality. And some scholars study the interpretation of the characteristic expression of the hidden space variables of the model. The common generative models are VAE(Variational auto-encoder) \cite{DBLP:journals/corr/KingmaW13} and GAN(Generative Adversarial Networks) \cite{DBLP:journals/corr/GoodfellowPMXWOCB14}. They aim to disentangle the features of hidden space variables and semantically separate the feature expression of the original model, so that different regions of hidden space variables represent different semantic information that human beings can understand, so as to achieve interpretability. The advantage of this method is that the hidden space variables can be controlled and the desired output results can be generated. The disadvantage is that if there is too much semantic information in the input, it is difficult to completely disentangle the hidden space variables. The above are some main interpretability research directions at present, but it can be seen that they are related to the internal structure of the model or the input and output of the model.

For most deep learning classification tasks, the process is to put the input into the neural network, and then get the predicted category, which is kind of unexplainable. Therefore, this paper proposes a multi granularity semantic recognition model, which can recognize the fine-grained semantics in a picture, and then make decisions and infer the coarse-grained categories according to these semantics. For the classification task, we believe that an interpretable output needs to correctly identify the fine-grained semantic information in the picture and correctly predict the coarse-grained classification categories. This paper also proposes a new index to evaluate the interpretability of the model. In addition, combined with the current advanced deep learning model, this paper obtains the baseline performance of the related interpretability index. 

The contributions of our work can be summarized as follows:
\begin{itemize}
\item We first propose a multi-semantic image recognition model, which enables human beings to understand the decision-making process of the neural network. 
\item We present an interpretable evaluation index, which has the advantage of testing whether the decision-making process of neural networks is in line with human thinking mode. We can quantitatively evaluate the interpretability of the model.
\item We propose our model and method, which can learn the semantic information that affects the results of image classification. Our method can help neural networks learn about human decision-making processes for image classification.
\end{itemize}

\section{Related Work}

It is a very popular research direction to explain the hidden space variables of the model. The representative generation models are VAE\cite{DBLP:journals/corr/KingmaW13} and GAN\cite{DBLP:journals/corr/GoodfellowPMXWOCB14}. VAE aims to disentangle the hidden space variables learned by the encoder so that human beings can control some specific semantics of the output in the decoder. For feature disentangling, $\beta $-VAE \cite{DBLP:conf/iclr/HigginsMPBGBML17} can enhance the disentangling ability of the model by multiplying the $\beta$ hyperparameter before the KL divergence term of the traditional VAE loss function. On this basis, the authors of $\beta$-VAE further explained $\beta$-VAE with the idea of information bottleneck, and explained why adding hyperparameters can enhance the disentangling ability\cite{burgess2018understanding}. Disentangling the representation space of GAN plays the same role. CGAN (conditional generic advantageous networks)\cite{DBLP:journals/corr/MirzaO14} model adds category labels to the discriminator and generator so that the learned feature expression space has semantic category information. InfoGAN model \cite{DBLP:conf/nips/ChenCDHSSA16} learns the disentangled representation of variables by combining GAN with mutual information.

LIME\cite{DBLP:conf/kdd/Ribeiro0G16} and SHAP\cite{DBLP:conf/nips/LundbergL17} treat the internal structure of the model as a black box, which is an interpretable method independent of the model. BayLIME \cite{DBLP:journals/corr/abs-2012-03058} model is based on LIME, and its parameters are automatically fitted from samples according to the Bayesian model or obtained from prior knowledge of the specific application. KernelSHAP \cite{DBLP:conf/nips/LundbergL17} model uses a linear model to simulate the input and output of the original neural network model, and the characteristics with strong discrimination will get larger coefficients in the linear model. The author also proposed the TreeSHAP \cite{DBLP:journals/corr/abs-1802-03888} model to explain tree-based models such as decision tree, random forest, and gradient lifting tree.

Among the interpretability methods related to the internal structure of the model, the heat map is a common visualization method, which aims to find the characteristics strongly related to the prediction results of the model. Occlude an area of the input image and observe the impact of the occluded area on the model prediction results\cite{zeiler2014visualizing}. This paper also proposes a deconvolution method to map the specific neurons of a certain layer back to the pixel space to get the picture with the same size as the input and observe which part of the semantic features of the input are learned. CAM(Class Activation Mapping) \cite{DBLP:conf/cvpr/PopeKRMH19} series is also a method of visualization. CAM is the weighted linear sum of multi-layer feature maps, and the activation map of a specific category is upsampled to the same size as the input image so that the image region most relevant to a specific category can be intuitively identified.

\subsection{COCO Dataset}
Datasets play an important role in the development of deep learning. COCO dataset is a large dataset that can be applied to multiple tasks such as multi-label classification, target detection, keypoint detection, semantic segmentation, and instance segmentation. Its purpose is to put the target recognition problem in the context of a broader scene understanding problem, so as to promote the development of target recognition. Its images are collected in daily complex scenes. The COCO dataset \cite{lin2014microsoft} includes 91 public object categories, 328000 pictures, and 2500000 labeled instances. It is the largest dataset that can be applied to semantic segmentation tasks so far. There are 80 categories and more than 330000 pictures provided in semantic segmentation tasks, of which 200000 are marked, and the number of individuals in the whole dataset exceeds 1.5 million. At the pixel level, it has been labeled with different semantic labels at different levels of granularity. It has 80 kinds of fine-grained semantics, and some particular fine-grained semantics belong to a particular coarse-grained semantics, with a total of 12 kinds of coarse-grained semantics. In the COCO2017 dataset, there are 118287 pictures in the training set, 5000 pictures in the verification set, and no public test set. In the challenge of ICCV2019 - COCO + Mapillary Joint Recognition Challenge Workshop, the COCO dataset provides four competition tasks: instance segmentation, panoramic segmentation, keypoint detection, and densepose, and the winner will report on ICCV2019 Workshop. The latest COCO competition is held by the  COCO-LVIS Joint Workshop of the ECCV 2020 Conference, which also includes target detection, instance segmentation, human keypoint detection, panoramic segmentation, and other tracks. It can be said that the COCO challenge has become the most influential target recognition competition in the field of artificial intelligence, with high recognition and popularity in both academia and industry.

\section{Dataset and Interpretability Evaluation Index}
In this part, we introduce our dataset annotation and interpretability evaluation index.

\subsection{Dataset Annotation}
COCO dataset is a large dataset that can be applied to multiple tasks such as target detection and semantic segmentation. It has been labeled with semantic labels at the pixel level and defined its specific semantic attribute names. Its fine-grained semantics include person, bicycle, car, motorbike, airplane, bus, train, truck, boat, traffic light, fire hydrant, stop sign, parking meter, bench, bird, cat, dog, horse, sheep, cow, elephant, bear, zebra, giraffe, backpack, umbrella, handbag, tie, suit, frisbee, ski, snowboard, sports ball, kit, baseball bat, baseball glove, skateboard, surfboard, tennis racket, bottle, wine glass, cup, fork, knife, spoon, bowl, banana, apple, sandwich, orange, broccoli, carrot, hotdog, pizza, donut, cake, chair, sofa, potted plant, bed, dining table, toilet, television, laptop, mouse, remote, keyboard, cellphone, microwave, oven, sink, refrigerator, book, clock, vase, scissors, teddy bear, hairdryer and toothbrush, with a total of 80 classes. Some specific fine-grained semantics belong to specific coarse-grained semantics. The coarse-grained semantics include person, vehicle, outdoor, animal, accessory, sports, kitchen, food, furniture, electronic, appliance, and indoor, with a total of 12 classes. The COCO dataset can also be used for the task of multi-label classification, that is, to predict the fine-grained semantics in a picture. Our idea is to improve the coco dataset to make it suitable for our multi-granularity semantic recognition model, so that it not only has two-level granularity labels of coarse-grained and fine-grained at the pixel level but also at the image level, that is, for a picture, it contains both fine-grained semantic label information in the image and coarse-grained category label information of the whole picture.

To achieve this goal, we first need to filter the original COCO dataset, because, at the pixel level of a picture, it can contain fine-grained semantics under different coarse-grained semantics, which makes it impossible to label the category label of the whole picture. Our rule is to make all the fine-grained semantic information in a picture belong to the same coarse-grained semantics, and this coarse-grained semantics is the category label we want to annotate the whole picture, so that the dataset has two-level granularity labels of coarse-grained and fine-grained. Apply our rules to the COCO2017 training set and validation set to get the filtered images.

The second is to annotate the semantic labels of the dataset. We retained 80 classes of fine-grained semantic labels of the original COCO dataset and modified the original 12 classes of coarse-grained semantics. Both the fine-grained and coarse-grained semantics in the original COCO dataset contains the semantics of the person. We remove it from the coarse-grained semantics so that the semantics of the person in the fine-grained semantics become the public fine-grained semantics. It does not belong to a specific coarse-grained but can exist in each picture so that the number of coarse-grained semantics becomes 11. The attribute names of coarse-grained semantics are modified to make them more in line with the actual situation. The modified attribute names are transport, outdoor sign, animal, accessory, sports, dining tools, furniture decoration, indoor supply, electronic, appliance, and food.

Finally, there are 60554 pictures in the improved dataset, including 80 fine-grained semantic labels and 11 coarse-grained category labels. There are 58107 training sets and 2447 test sets. The hierarchy of the dataset is shown in Figure 1.

%\begin{figure*}
%	\centering
%	\includegraphics[width=2\columnwidth]{layer_tree.jpg}
%	\caption{Datasets are divided into coarse-grained categories and fine-grained semantics. %Each specific coarse-grained category has its specific fine-grained semantics.}
%	\label{fig:framework}
%\end{figure*}

\subsection{Interpretability Evaluation Index}
We think that in order to make the output results of the neural network interpretable, the output results need not only to correctly predict the coarse-grained category labels of the picture but also to correctly identify all the fine-grained semantics in a picture. The neural network needs to first identify the fine-grained semantics in the picture, and then judge which coarse-grained category the whole picture belongs to according to these semantics. This decision-making process is more in line with the human thinking mode and has a hierarchy. Our interpretable decision path is shown in Figure 2.

\begin{figure*}[htbp]
    \centering    %居中
    \subfigure[The decision-making process of unexplainable neural networks on traditional datasets] %第一张子图
    {
    \begin{minipage}[t]{\linewidth}
    \centering
        \includegraphics[width=5in]{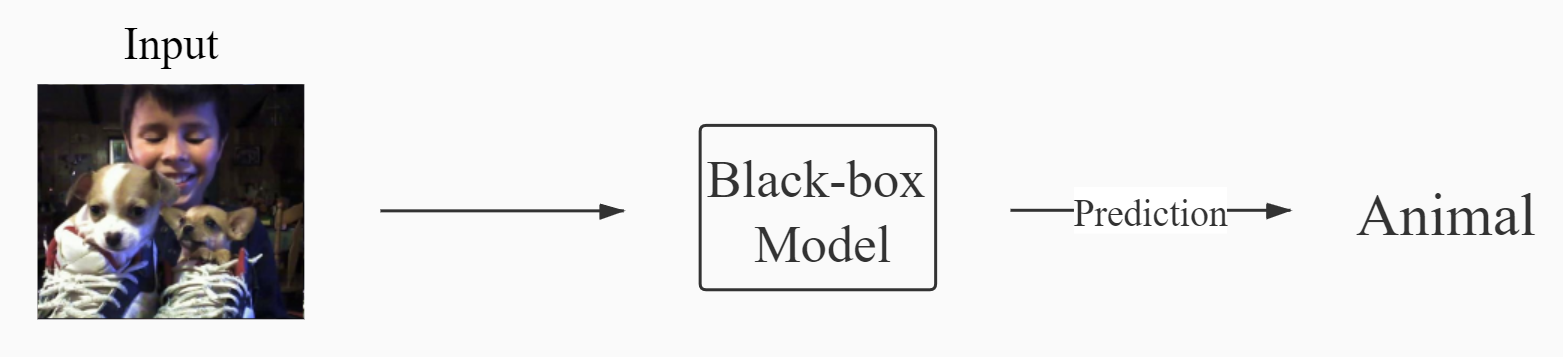}
    \end{minipage}
    }
    
    \subfigure[The decision-making process of interpretable neural networks on our datasets] %第二张子图
    {
    \begin{minipage}[t]{\linewidth}
    \centering      %子图居中
        \includegraphics[width=7in]{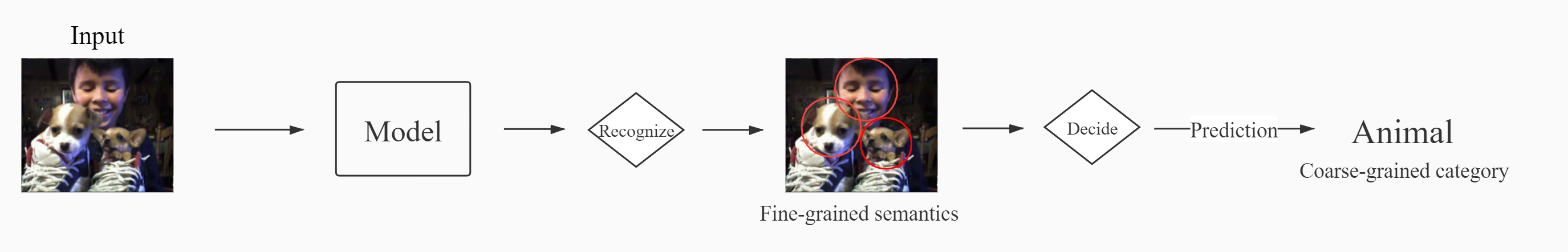} 
    \end{minipage}
    }
    \caption{(a) In the traditional image classification dataset, after the input is put into the neural network, the black-box model obtains the final output result according to its decision-making process, which is unknown to human beings. (b) On our model, after the input is also put into the neural network, the model will first recognize the fine-grained semantics in the picture, and then make decisions to obtain the final coarse-grained category results according to these semantics, which is in line with human decision-making thinking mode. In this example, the neural network first recognizes the fine-grained semantics of person and dog, then infers that the picture belongs to the coarse-grained category of the animal according to the two semantics.} %  %大图名称
    \label{fig:framework}  %图片引用标记
\end{figure*}
 
Therefore, we define the parameter FTCT (Fine-grained True Coarse-grained True), that is, it is considered as an interpretable prediction only when the prediction of both fine-grained semantics and coarse-grained categories is correct. Accordingly, three parameters FTCF (Fine-grained True Coarse-grained False), FFCT (Fine-grained False Coarse-grained True), and FFCF (Fine-grained False Coarse-grained False) are also defined. The four parameters can be represented in Table 1.

\begin{table}[!htbp]
    \resizebox{.95\columnwidth}{!}{%对字体大小调整，避免表格太长{
    \centering
    \begin{tabular}{|c|c|c|}
    \hline
    \diagbox{Fine-grained prediction}{Coarse-grained prediction}&True&False\\ %添加斜线表头
    \hline
    True&FTCT&FTCF\\
    \hline
    False&FFCT&FFCF\\
    \hline
    \end{tabular}
    }
    \caption{Interpretability parameter definition.}
    \label{table}
\end{table}
%\begin{center}
%\caption{Table 1.Interpretability parameter definition}
%\end{center}

\par In our opinion, we think that there are four interpretability evaluation indexes: IA (Interpretability Accuracy), IR (Interpretability Recall), IP (Interpretability Precision), and IF (Interpretability F-score). Their formulas are shown as follows.

IA is defined as the proportion of samples whose coarse-grained and fine-grained information are both correctly predicted by the model in all samples.
\begin{equation} \label{LinearFunction1}
    IA=\frac{FTCT}{FTCT+FTCF+FFCT+FFCF}%FTCT/(FTCT+FTCF+FFCT+FFCF)
\end{equation}
\par IR is defined as the proportion of samples whose coarse-grained and fine-grained information are both correctly predicted by the model in the samples whose coarse-grained category information is correctly predicted by the model.
\begin{equation} \label{LinearFunction2}
    IR=\frac{FTCT}{FTCT+FFCT}%FTCT/(FTCT+FFCT)
\end{equation}
\par IP is defined as the proportion of samples whose coarse-grained and fine-grained information are both correctly predicted by the model in the samples whose fine-grained semantic information is correctly predicted by the model.
\begin{equation} \label{LinearFunction3}
    IP=\frac{FTCT}{FTCT+FTCF}%FTCT/(FTCT+FTCF)
\end{equation}
\par It is defined as a combination of IR and IP.
\begin{equation} \label{LinearFunction4}
    IF=\frac{(\alpha^2+1)\cdot IP \cdot IR}{IP+IR}%(\alpha^2+1)\cdot IP \cdot IR/(IP+IR)
\end{equation}
In the experiment, you can set the $\alpha$ value to 1, which is the IF1 value.

\section{Experiments}
In this part, we first introduce our model methods applied to our dataset. Further, combined with the current advanced feature extractors such as VGG, ResNet, and DenseNet\cite{2016Densely}, we carried out experiments on our dataset with our model method and obtained the baseline results of related interpretability evaluation indexes.

\subsection{Model Method}
According to the characteristics of our dataset, combined with the feature extractor of various current CNN models, our model method is to train a two-level label classifier after the feature extractor to fit coarse-grained and fine-grained two-level labels respectively. The number of neurons in the final output layer of the first-level label classifier is equal to the number of fine-grained semantics, that is, let each neuron learn a specific fine-grained semantic information, which is a multi-label and multi-classification task. Suppose the number of samples is N and the number of neurons is T. According to the selection rules of the number of fine-grained semantics by the subsequent first-level label classifier, there are two loss functions:

1) When the activation function adopts softmax function, its loss function is:
\begin{equation} \label{LinearFunction5}
    L_{fine}=-\sum_{i=1}^{N}\sum_{t=1}^{T}\alpha logp_t(X_i)(1-p_t(X_i))^\gamma y_t(X_i)
\end{equation}
where $\alpha$ is the weight of each fine-grained semantics in the sample, and the $\gamma$ is the super parameter, which is to reduce the weight of easily classified samples, to make the model more focused on the samples that are difficult to be classified during training\cite{lin2017FOCAL}.

2) When the activation function adopts sigmoid function, its loss function is:
\begin{small}
\begin{equation} \label{LinearFunction6}
    L_{fine}=-\sum_{i=1}^{N}\sum_{t=1}^{T}y_t(X_i)logp_t(X_i)+(1-y_t(X_i))log(1-p_t(X_i))
\end{equation}
\end{small}
The number of neurons in the final output layer of the second-level label classifier is equal to the number of coarse-grained categories so that each neuron corresponds to a specific coarse-grained category. Suppose the number of samples is N and the number of neurons is M. The loss function can be expressed as:
\begin{equation} \label{LinearFunction7}
    L_{coarse}=-\sum_{i=1}^{N}\sum_{m=1}^{M}y_m(X_i)logp_m(X_i)
\end{equation}

Therefore, the loss function of the whole model is:
\begin{equation} \label{LinearFunction8}
    L_{all}=L_{fine}+L_{coarse}
\end{equation}
Our model diagram is shown in Figure 2.
\begin{figure*}
	\centering
	\includegraphics[width=2\columnwidth]{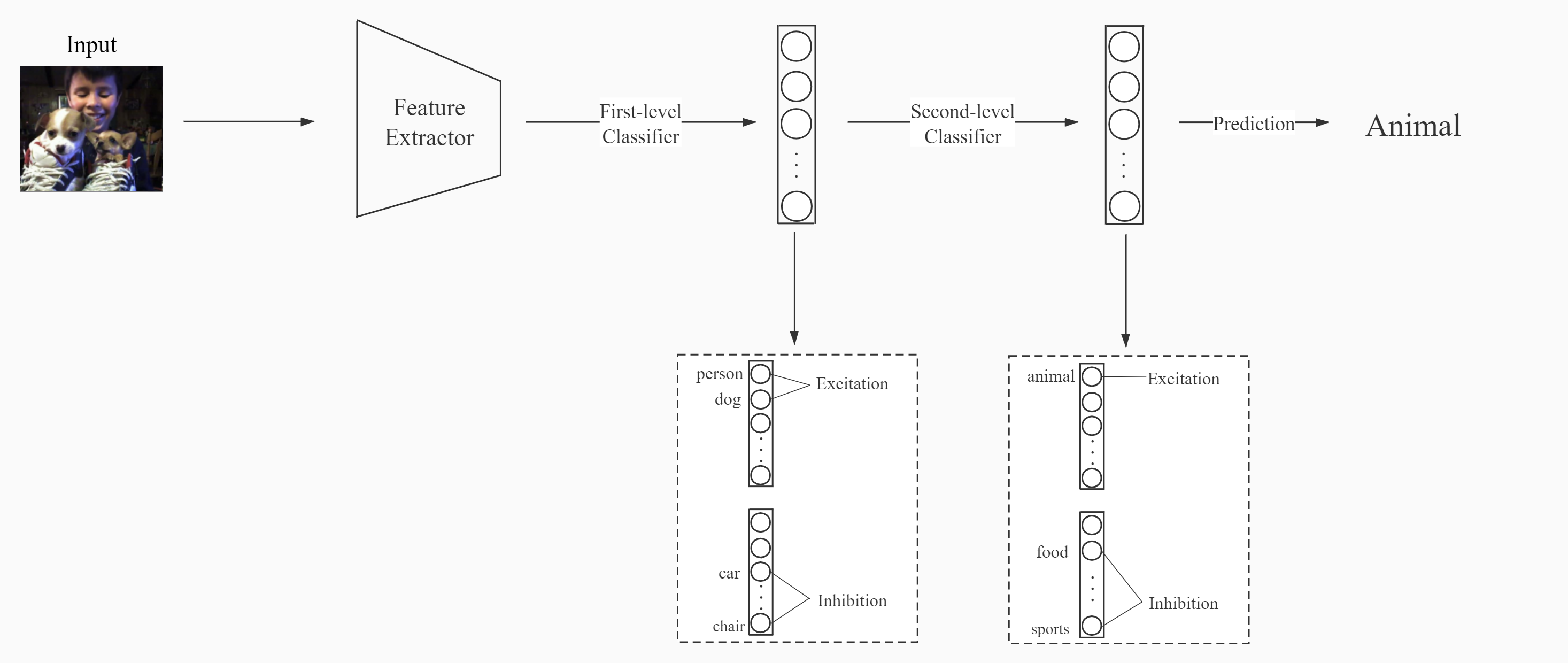}
	\caption{Each dimension of the final layer of the first-level classifier of our model corresponds to a specific fine-grained semantic information, and the final output of the second-level classifier is the predicted coarse-grained category. For the input picture as shown in the figure, the first-level classifier of the neural network needs to recognize the two fine-grained semantic information of person and dog, and then the second-level classifier judges and predicts that it is an animal category according to the recognized semantic information.}
 	\label{fig:framework1}
\end{figure*}

\subsection{Baseline Results of Interpretability Evaluation Index}
Firstly, we use the advanced CNN model to conduct experiments on our dataset, and only the coarse-grained categories are constrained by loss functions, so as to obtain the traditional unexplainable accuracy as a classification evaluation index. The experimental results are shown in Table 2.
\begin{table}[!htbp]
\resizebox{.95\columnwidth}{!}{%对字体大小调整，避免表格太长{
\begin{tabular}{ccc}
\hline
Model& Unexplainable Accuracy\\
\hline
VGG16& 0.8457\\
ResNet34& 0.8655\\
DenseNet121& 0.8946\\
\hline
\end{tabular}
}
\caption{Unexplainable accuracy.}
\end{table}
\par Then we retain the feature extractor part of the traditional CNN classic model, on this basis, we modify the full connection layer according to our method, that is, we use loss function to constrain both fine-grained semantics and coarse-grained categories, and obtain the relevant interpretability classification evaluation index we proposed.

Because the fine-grained semantics of the prediction output of the first-level classifier of the neural network needs to be compared with the actual fine-grained semantic labels, it involves the problem of selecting which semantics in the prediction output. The idea is to solve it in the following two situations.

1) The fine-grained semantic value of the prediction output can be passed through a softmax function, and the previous dimension (the same number as the real fine-grained semantic label) from large to small is taken as the output, and these are considered to be the semantics learned by the neural network. At this time, the loss function of the first-level classifier is formula 5. The experimental results are shown in Table 3.
\begin{table}[!htbp]
\centering
\resizebox{.95\columnwidth}{!}{%对字体大小调整，避免表格太长
\begin{tabular}{ccccc}
\hline
Model& IA &IP &IR &IF1\\
\hline
VGG16+Ours&0.6502&0.9947&0.7544&0.8586\\
ResNet34+Ours&0.6841&0.9981&0.7774&0.8715\\
DenseNet121+Ours&0.7295&0.9988&0.8102&0.8913\\
\hline
\end{tabular}
}
\caption{Interpretability index results when the focal loss is used as the loss function of the first level classifier.}
\end{table}

2) The fine-grained semantic value of the prediction output can be passed through a sigmoid function, and a threshold value can be set for the dimension. When the value is higher than the threshold value, it is considered that the neural network has learned semantics. Not vice versa. At this time, the loss function of the first-level classifier is formula 6. The threshold of the current experiment is 0.7. The experimental results are shown in Table 4.
\begin{table}[!htbp]
\centering
\resizebox{.95\columnwidth}{!}{%对字体大小调整，避免表格太长
\begin{tabular}{ccccc}
\hline
Model& IA &IP &IR &IF1\\
\hline
VGG16+Ours& 0.4947 & 1.0000&0.6146&0.7567\\
ResNet34+Ours& 0.5856 & 1.0000 &0.6799&0.8044\\
DenseNet121+Ours&0.5488 &1.0000 &0.6632&0.7900\\
\hline
\end{tabular}
}
\caption{Interpretability index results when the cross-entropy loss is used as the loss function of the first level classifier.}
\end{table}

After comparing the experimental results in several tables, although the experimental results of the traditional accuracy index look very high, it can be seen from the experimental results of IA and IR that there are some samples in which the neural network correctly predicts the coarse-grained categories, but the neural network does not recognize the correct fine-grained semantic information in these samples. This also reflects the black-box nature of the neural network, which obtains the output results in a way of thinking that human beings cannot understand. From the experimental results of IP, it can be seen that once the neural network correctly recognizes the fine-grained semantic information in the picture, it almost determines the coarse-grained category of the picture. This also reflects the superiority of our interpretability evaluation index. Our model method can also make the neural network learn more according to the human way of thinking, so as to realize the interpretability of the neural network.
% An example of a floating figure using the graphicx package.

\section{Conclusion}
In this work, we propose a multi-semantic image recognition model, which enables human beings to understand the decision-making process of the neural network. Besides, we propose the interpretability evaluation index, which can quantitatively evaluate the interpretability of the model. We aim to exploit human thinking and decision-making logic to evaluate the performance and explanation of neural networks. Further, we propose our model method, which can investigate the semantic information that affects the results of image classification. We attempt to make the neural network learn the human way of thinking, and make its output process conform to the human process of decision-making. Last, combined with the current advanced deep learning model, we exhibit the baseline performance of the relevant interpretability index of our model method.


%



%\bibliographystyle{IEEEtran}

\input{Main.bbl}

\bibliographystyle{IEEEtran}
%\bibliography{ref}
%\printbibliography

\end{document}